\documentclass[sigconf, nonacm]{acmart}

\usepackage{algorithm}
\usepackage{multirow}
\usepackage{algpseudocode}
\usepackage{tikz}
\usepackage{standalone}
\usepackage{listings}
\usepackage{tikzscale}
\usepackage{stfloats}
\usepackage{caption}
\usepackage{placeins}
\usetikzlibrary{shapes,positioning,fit,backgrounds,shadows}
\newcommand\vldbdoi{XX.XX/XXX.XX}
\newcommand\vldbpages{XXX-XXX}
\newcommand\vldbvolume{14}
\newcommand\vldbissue{1}
\newcommand\vldbyear{2020}
\newcommand\vldbauthors{\authors}
\newcommand\vldbtitle{\shorttitle} 
\newcommand\vldbavailabilityurl{URL_TO_YOUR_ARTIFACTS}
\newcommand\vldbpagestyle{plain}

\begin{document}
\title{SAVeD: Semantically Aware Version Discovery}

\author{Artem Frenk}
\orcid{0009-0007-0871-5691}
\affiliation{%
  \institution{Worcester Polytechnic Institute}
  \streetaddress{100 Institute Rd.}
  \city{Worcester}
  \state{MA}
  \postcode{01605}
}
\email{afrenk@wpi.edu}

\author{Roee Shraga}
\affiliation{%
  \institution{Worcester Polytechnic Institute}
  \streetaddress{100 Institute Rd.}
  \city{Worcester}
  \state{MA}
}
\email{rshraga@wpi.edu}

\begin{abstract}
Our work introduces SAVeD (Semantically Aware Version Detection), a contrastive learning-based framework for identifying versions of structured datasets without relying on metadata, labels, or integration-based assumptions. SAVeD addresses a common challenge in data science of repeated labor due to a difficulty of similar work or transformations on datasets.
SAVeD employs a modified SimCLR pipeline, generating augmented table views through random transformations (e.g., row deletion, encoding perturbations). These views are embedded via a custom transformer encoder and contrasted in latent space to optimize semantic similarity. Our model learns to minimize distances between augmented views of the same dataset and maximize those between unrelated tables. We evaluate performance using validation accuracy and separation, defined respectively as the proportion of correctly classified version/non-version pairs on a hold-out set, and the difference between average similarities of versioned and non-versioned tables (defined by a benchmark, and not provided to the model).
Our experiments span five canonical datasets from the Semantic Versioning in Databases Benchmark, and demonstrate substantial gains post-training. SAVeD achieves significantly higher accuracy on completely unseen tables in, and a significant boost in separation scores, confirming its capability to distinguish semantically altered versions. Compared to untrained baselines and prior state-of-the-art dataset-discovery methods like Starmie, our custom encoder achieves competitive or superior results.
\end{abstract}
\maketitle

\pagestyle{\vldbpagestyle}
\begingroup\small\noindent\raggedright\textbf{PVLDB Reference Format:}\\
\vldbauthors. \vldbtitle. PVLDB, \vldbvolume(\vldbissue): \vldbpages, \vldbyear.\\
\href{https://doi.org/\vldbdoi}{doi:\vldbdoi}
\endgroup
\begingroup
\renewcommand\thefootnote{}\footnote{\noindent
This work is licensed under the Creative Commons BY-NC-ND 4.0 International License. Visit \url{https://creativecommons.org/licenses/by-nc-nd/4.0/} to view a copy of this license. For any use beyond those covered by this license, obtain permission by emailing \href{mailto:info@vldb.org}{info@vldb.org}. Copyright is held by the owner/author(s). Publication rights licensed to the VLDB Endowment. \\
\raggedright Proceedings of the VLDB Endowment, Vol. \vldbvolume, No. \vldbissue\ %
ISSN 2150-8097. \\
\href{https://doi.org/\vldbdoi}{doi:\vldbdoi} \\
}\addtocounter{footnote}{-1}\endgroup

\ifdefempty{\vldbavailabilityurl}{}{
\vspace{.3cm}
\begingroup\small\noindent\raggedright\textbf{PVLDB Artifact Availability:}\\
The source code, data, and/or other artifacts have been made available at \url{https://github.com/afrenkai/Semantic-Aware-Version-Discovery}.
\endgroup
}

\section{Introduction}
In any field revolving around data, it is reasonable to assume that outside of a heavily controlled environment, data is "unclean" and "unorganized", that is, it lacks structure. In both corporate and academic settings, datasets often require some transformation in order to be of use to an engineer or developer. However, the transformations performed on the same piece of data vary wildly, even on a set of data considered to be "solved". \cite{CaiZhu2015}


Take, for example, the well known "Titanic" dataset, used as a classification benchmark for many years.\footnote{https://www.kaggle.com/competitions/titanic} The goal of this dataset is to predict whether a passenger survived the Titanic disaster based on features such as age, class, and fare. When preparing this data for machine learning, one developer may choose to one-hot encode categorical variables (such as passenger class or embarkation port) to create binary indicator features, while another may prefer to use learned embeddings to capture semantic relationships between categories. These are all design decisions made to optimize model performance, and each results in a distinct \textit{version} of the dataset. In the former example, the resulting table has 1733 columns (using scikit-learn's OneHotEncoder method). In the latter case, the table has merely 167 columns. Examples of these transformations can be found in our repository\footnote{\url{https://github.com/afrenkai/Semantic-Aware-Version-Discovery}}. These tables are both derived purely from the internal data of the original table, and as such constitute valid versions. 

Discovering these versions is crucial for avoiding redundant work and leveraging existing data science efforts. The discovery of these valid tables is called \textit{Version Discovery}. This problem represents the first explicit exploration of automated version discovery \cite{10.14778/3583140.3583169}, addressing a gap in prior work on data versioning and management. 

Version discovery can be viewed as a dataset discovery task. It exists within the broader landscape of dataset discovery methods such as keyword search \cite{4430467} and query search \cite{DBLP:journals/corr/GaoC17}. More recently, dataset discovery has focused on the tasks of Table Union Search \cite{10.14778/3192965.3192973} and Joinable table search \cite{10.1145/3299869.3300065}, both of which focus on the interaction between tables in a data lake. In the realm of version discovery, prior work has focused on creating benchmarks of table versions using seed tables \cite{10.14778/3583140.3583169} and creating benchmarks using Large Language Models (LLMs) \cite{10.1145/3627673.3679157}. However, no work has been performed (to the authors' knowledge) on version discovery itself, that is, finding whether existing modifications have been made to a given dataset already. For the task of table union search, contrastive learning \cite{fan2023semanticsawaredatasetdiscoverydata} has been used as a means of augmenting the search in an efficient manner. We posit that contrastive learning on table pairs can also be used as a means of augmentation to discover realistic, internal-driven transformations for tables in a data lake, creating a framework for semantic-aware version discovery.

We make the following specific contributions:
\begin{itemize}
    \item We formalize the problem of semantic version discovery and provide a theoretical framework defining version relationships through semantics-preserving transformations.
    \item We introduce SAVeD, a contrastive learning-based framework that learns to identify table versions without requiring metadata, labels, or prior knowledge of version relationships.
    \item We design a suite of table-specific augmentations that preserve semantic content while creating diverse structural representations for effective contrastive learning.
    \item We demonstrate through extensive experiments on the Semantic Versioning in Databases Benchmark that SAVeD achieves superior performance compared to untrained baselines and competitive results against state-of-the-art methods like Starmie, while operating in a fully unsupervised setting.
\end{itemize}


\section{Related Work}
\subsection{Data Versioning}
Data versioning has been traditionally approached from a metadata-driven perspective. The foundational work of Sciore \cite{Sciore94} established theoretical frameworks for database versioning, while Bhardwaj, et. al's work on DataHub \cite{bhardwaj2014datahubcollaborativedatascience} identified key challenges in collaborative data science environments. Explain Da V\cite{10.14778/3583140.3583169} introduced the concept of semantic versioning in databases, establishing benchmarks for version identification but assuming prior knowledge of version relationships.

Modern version control systems like  
DVC \cite{ruslan_kuprieiev_2025_17037377}, Pachyderm \cite{Pachyderm}, and even AutoML plaforms like MLflow\cite{Zaharia2018AcceleratingTM} provide infrastructure for tracking data changes but require explicit version declarations and metadata maintenance. Git \cite{git} extends traditional version control to data but struggles with large binary files, relying on approaches like its Large File System (LFS). Recent approaches like Huggingface's Xet have bypassed this limitation, but inherently lack semantic understanding.

The database community has explored temporal databases \cite{404027} and schema evolution \cite{inproceedingsmanousis} extensively, but these approaches focus on structural changes rather than semantic relationships. 
Work by Fox et al.\cite{10.1145/3627673.3679157} explored LLM-based approaches for generating synthetic table versions using prompt engineering and GPT models, demonstrating the complexity of realistic version creation but focusing on benchmark generation rather than discovery. Shraga and Miller \cite{10.14778/3583140.3583169} developed Explain-Da-V and established the Semantic Versioning in Databases Benchmark (SDVB) for explaining dataset changes but assumed known version relationships.

Unlike these approaches, SAVeD operates without metadata or labeled relationships, learning version semantics directly from data structure and content through self-supervised contrastive learning.
\subsection{Data Discovery}

Table discovery and search have been extensively studied in database research, evolving from early schema matching approaches to modern semantic-aware systems. Traditional approaches like COMA \cite{inproceedings} and Similarity Flooding \cite{10.5555/876875.879024} focused on structural and linguistic similarity.

Web table discovery systems emerged with the growth of HTML tables. WebTables \cite{10.14778/1453856.1453916} and subsequent work by Lehmberg et al.\cite{10.1145/2872518.2889386} demonstrated large-scale table extraction and matching from web sources. However, these approaches primarily address structural heterogeneity rather than semantic versioning relationships.

Recent advances in deep learning leverage deep semantic relationships for table understanding. TaBERT\cite{yin2020tabertpretrainingjointunderstanding} and TAPAS\cite{Herzig_2020} learn joint representations of tables and natural language, enabling question-answering over tabular data. Starmie\cite{fan2023semanticsawaredatasetdiscoverydata} uses pre-trained RoBERTa models for table classification and retrieval tasks, achieving state-of-the-art performance on benchmark datasets like the aforementioned Semantic Versioning in Databases dataset. 

The advent of foundation models has spawned new approaches: TAPEX \cite{liu2022tapextablepretraininglearning} learns to execute SQL queries by treating tables as structured text and the ToTTo \cite{parikh2020tottocontrolledtabletotextgeneration} benchmark addresses table-to-text generation. While these approaches excel at table comprehension tasks, they require structured supervision and operate fundamentally differently from version discovery, which needs to learn semantic relationships from unlabeled table pairs without assuming well-defined task objectives.

The subtask of joinability detection represents a related but distinct problem. JOSIE \cite{10.1145/3299869.3300065} and LSH Forest \cite{10.1145/1060745.1060840} focus on finding tables that can be joined based on overlapping values, while our work addresses semantic equivalence despite structural differences. Starmie \cite{fan2023semanticsawaredatasetdiscoverydata} represents the most closely related work to our approach. It uses pre-trained RoBERTa models with contrastive learning for table understanding tasks, achieving state-of-the-art results on various benchmarks including the Semantic Versioning in Databases dataset. However, Starmie focuses on general table discovery and requires supervised fine-tuning for specific tasks. In contrast, SAVeD is specifically designed for version discovery through unsupervised contrastive learning, with custom augmentations tailored to capture version relationships without labeled data. For a detailed discussion of contrastive learning methods in data integration and discovery, see Section 4.1.
\cite{fan2023semanticsawaredatasetdiscoverydata} has used contrastive learning along with labeled pairs to learn a semantic-aware representation of tables for the task of joinability. While the methodology is solid, Starmie’s contrastive objective is anchored in supervised pairs curated for joinability and table discovery, As a result, its embedding space is shaped to capture cross-table semantic similarity rather than intra-dataset evolutionary changes. Second, the augmentations used in Starmie are designed to enhance robustness to surface-level noise (e.g., column permutations or name variations), but they do not simulate version-like transformations such as incremental schema evolution, value drift, or attribute-level refinements. These differences cause Starmie’s embeddings to collapse version-specific nuances that are essential for version discovery. By contrast, SAVeD’s unsupervised pipeline forces the model to learn  the kinds of fine-grained relationships that supervised table-discovery pre-training tends to overlook. 
\section{Version Discovery}
\subsection{What is a version?}
Data scientists commonly create multiple derived variants of the same underlying dataset by applying preprocessing, encoding, feature engineering, or sampling operations. We call two tables versions of each other when one can be obtained from the other via a sequence of semantics-preserving transformations (for example, column encoding changes, benign value normalizations, or row reordering).

Importantly, in this work we define versions as being derived purely from the table's internal content and metadata. By "internal metadata", we mean information that can be computed from the table itself (schema, column names, data types, value distributions, summary statistics, and column-level histograms), as opposed to external annotations, labels, web-derived metadata, or integrations with other datasets. The family of allowed transformations $\mathcal{P}$ therefore contains only operations that act on $T$ (its tuples and attributes) without consulting outside sources. This distinction separates version discovery from tasks that rely on external enrichment or ground-truth labels.

\subsection{Approach}
Our approach extends contrastive learning to the novel domain of table version discovery, where the challenge lies in designing augmentations that preserve semantic content while creating sufficient structural diversity for effective learning.
Prior work in semantic data versioning has assumed that a tables $T$ and $T'$ are versions of each other, but an explicit definition for what a \textit{version} constitutes has never been defined. 
Let a dataset be denoted by a table $T$ , composed of a set of attributes $T_A =\left \{ A_1, \dots A_n \right\}$ and tuples $T_r = \left \{r_1 \dots r_m \right \}$. Let each tuple defined by $r_i = \langle ri_0, ri_1 ... ri_n \rangle$, such that $r_{i0}$ can be easily recognized as the tuple identifier and $r_{ij} (j \neq 0)$ can represent a value assigned to the attribute $A_j$ in the tuple $r_i$. 
With that definition, let $T$ and $T'$ be two datasets. We say that $T'$ is a version of $T$ if there exists a transformation $p$ such that $p(T) = T'$, where $p$ belongs to a family of semantics-preserving transformations $\mathcal{P}$.


The version relationship is then defined as:
$$\text{Version}(T, T') \iff \exists p \in \mathcal{P}^* : p(T) = T'$$
where $\mathcal{P}^*$ denotes the closure of $\mathcal{P}$ under composition.

In practice, exact equality is replaced by semantic similarity: $p(T) \approx T'$ if $\theta(p(T), T') \geq \xi$ using similarity measure $\theta: \mathcal{T} \times \mathcal{T} \rightarrow [0,1]$ and threshold $\xi \in (0,1]$.

\textbf{Theoretical Properties:} The version relation exhibits the following properties:
\begin{itemize}
    \item \textit{Reflexivity:} $\text{Version}(T, T)$ (identity transformation)
    \item \textit{Symmetry:} $\text{Version}(T, T') \Rightarrow \text{Version}(T', T)$ (for invertible $p$)
    \item \textit{Transitivity:} $\text{Version}(T, T') \land \text{Version}(T', T'') \Rightarrow \text{Version}(T, T'')$
\end{itemize}

This defines a tolerance relation \cite{zeeman1962topology} on $\mathcal{T}$, which is weaker than an equivalence relation due to the approximate nature of $\theta$.  
\subsection{Motivating Example}
To illustrate the problem, consider, again, the Titanic dataset. Table~\ref{tab:basetitanic} shows the original (base) table. Analysts often produce derived versions of this dataset: for example, Table~\ref{tab:ohe_titanic} shows a one-hot encoded variant with many binary indicator columns created from categorical attributes, while Table~\ref{tab:embed_titanic} shows a version where categorical fields are replaced by learned embedding vectors and name columns are hashed to integer identifiers. Both derived tables preserve the semantic content of the base table (passenger records, classes, fares, survival labels) but differ substantially in column layout and dimensionality.

This example demonstrates the core challenge of version discovery: automatic methods must be invariant to superficial structural changes (encoding, ordering, column addition/removal) while sensitive to semantic changes that alter the meaning of the data. The goal of version discovery is to provide an unsupervised solution to this issue, and as opposed to related table search \cite{10.1145/2213836.2213962}, version semantics need to be explicitly captured. All prior exploration in Data Versioning has proceeded under the assumption that the relationship between tables is known beforehand, but this must be discarded beforehand in order to best learn a relationship between $T$ and $T'$, taking into account the lack of labeled examples, and the motivations behind the problem.


\section{SAVeD: Semantic Aware Version Discovery}
\subsection{Preliminaries: Contrastive Learning}

Contrastive learning has emerged as a fundamental paradigm for self-supervised representation learning, with theoretical foundations rooted in metric learning \cite{1640964} and noise contrastive estimation \cite{JMLR:v13:gutmann12a}. The core principle involves learning representations that pull semantically similar samples together while pushing dissimilar samples apart in the embedding space.

Early work in computer vision established the effectiveness of contrastive approaches. Deep Metric Learning \cite{sym11091066} and Triplet Loss\cite{Schroff2015FaceNetAU} demonstrated that carefully constructed positive and negative pairs could learn discriminative embeddings. The breakthrough came with SimCLR\cite{chen2020simpleframeworkcontrastivelearning}, which showed that simple data augmentations combined with large batch sizes and strong data augmentation on the fly could achieve state-of-the-art self-supervised performance, rivaling supervised methods on ImageNet.

Subsequent developments expanded contrastive learning across domains. In natural language processing, SimCSE \cite{gao2022simcsesimplecontrastivelearning} applied contrastive learning to sentence embeddings, while ConSERT\cite{yan2021consertcontrastiveframeworkselfsupervised} and DiffCSE\cite{chuang2022diffcsedifferencebasedcontrastivelearning} refined the approach with improved augmentation strategies.
The theoretical understanding of contrastive learning has evolved significantly. Wang and Isola\cite{wang2022understandingcontrastiverepresentationlearning} provided theoretical analysis showing that contrastive learning implicitly performs spectral clustering. HaoChen et al. \cite{DBLP:journals/corr/abs-2106-04156} established generalization bounds, while Tosh et al. \cite{tosh2021contrastivelearningmultiviewredundancy} analyzed the optimization landscape of contrastive objectives.

Recent advances have focused on addressing key limitations: MoCo\cite{he2020momentumcontrastunsupervisedvisual} introduced momentum-based negative sampling to enable larger effective batch sizes, SwAV\cite{caron2021unsupervisedlearningvisualfeatures} eliminated the need for negative pairs through clustering, and BYOL\cite{grill2020bootstraplatentnewapproach} achieved contrastive learning without explicit negative samples.

In tabular data, applications have been limited. VIME\cite{NEURIPS2020_7d97667a} applied contrastive pretraining to tabular data for downstream prediction tasks, while TabNet\cite{arik2020tabnetattentiveinterpretabletabular} incorporated attention mechanisms but didn't leverage contrastive learning. More recent work by Bahri et al.\cite{bahri2022scarfselfsupervisedcontrastivelearning} introduced SCARF for tabular representation learning, but focused on supervised fine-tuning rather than discovery tasks. Most recently, Starmie \cite{fan2023semanticsawaredatasetdiscoverydata} leveraged contrastive learning for dataset discovery.
Our approach to version discovery centers around using contrastive learning techniques, which allow models a label-free way to distinguish between similar and dissimilar data points. We implement a SimCLR-style framework \cite{chen2020simpleframeworkcontrastivelearning}, generating positive and negative pairs from a given batch of samples, and is created in a way to address the challenges of locating semantically similar table versions ("versions" of a dataset). Since version discovery requires identifying different representations of the same data via transformation, supervised methods for the task of version discovery are impractical, due to the lack of labeled data for the task. 



\subsection{Augmentation} 
A necessary part of our framework is to generate two augmented views (positive pairs) of the same table. These views are, by design, semantically equivalent but structurally unique, which enable the model to learn invariant features that can capture latent information of the table.

We consider eight parameterized augmentation operations:

(1) \textbf{Random Column Dropout ($P_1$):} Randomly removes columns with probability $P_1$, simulating feature selection processes where analysts remove irrelevant or redundant attributes.

(2) \textbf{Random Dummy Encoding ($P_2$):} Converts categorical variables to binary indicator variables with probability $P_2$, emulating common preprocessing steps for machine learning pipelines.

(3) \textbf{Random Row Shuffling ($P_3$):} Reorders table rows with probability $P_3$, ensuring the model learns order-invariant representations while preserving semantic content.

(4) \textbf{Random One-Hot Encoding ($P_4$):} Applies one-hot encoding to categorical columns with probability $P_4$, representing another common data transformation approach.

(5) \textbf{Missing Value Injection ($P_5$):}
With a small (P = 0.02) per-cell probability, replaces a value with a \texttt{NaN} value to simulate corrupted or incomplete data, which is more realistic. and encourages SAVeD to learn a more robust imputation handling or other way to address missing values. 

(6) \textbf{Gaussian Jitter ($P_6$):} applies gaussian noise $z \sim \mathcal{N}(0, P_6)$ where $0<P_6\leq 0.01$ which simulates measurement noise and promotes more robustness in the training process. 

(7) \textbf{Column Order Shuffling ($P_7$):} Randomly permutes the column order of the table. This is useful in contexts where column order should not influence the model and helps test invariance to feature ordering.

(8) \textbf{Row Dropping ($P_8$):}
Randomly removes a fraction of the dataset rows. This operation simulates data loss or sub-sampling scenarios often encountered in real-world systems. 

\subsection {Tokenization}
\begin{figure}
    \includegraphics[width=1\linewidth]{simclr_blocks.tex}\\
    \caption{Table View Creation Pipeline}
    \label{fig:simclr_blocks}
\end{figure}
Given an input table $T$, we generate two views: $T_{ori}$ and $T_{aug}$ (with transformations applied according to the probability parameters). This approach ensures that positive pairs share semantic content while exhibiting sufficient structural diversity to prevent trivial solutions during contrastive learning.
Once $T_{ori}$ and $T_{aug}$ are generated, they undergo a specialized tokenization process designed for structured tabular data. Unlike traditional approaches that use pre-trained BERT tokenizers, we implement a custom tokenization strategy optimized for table content.

The augmentation probabilities are treated as hyperparameters and optimized during training to balance between preserving semantic information and creating meaningful structural variations.

\textbf{Table-to-Text Conversion:} Each table is converted to a string representation using the format:
$$\text{table\_str} = \texttt{COL\_col1 val1\_1 val1\_2 ... COL\_col2 val2\_1 ...}$$

This format preserves both structural information (column boundaries) and content while enabling standard Natural Language Processing (NLP) tokenization techniques.

\textbf{BPE Tokenizer:} We train a Byte-Pair Encoding (BPE) tokenizer specifically on tabular data with the following characteristics:
\begin{itemize}
    \item Vocabulary size: 12,000 tokens (optimized through hyperparameter search)
    \item Special tokens: \texttt{<UNK>}, \texttt{<PAD>}
    \item Minimum frequency threshold: 2 occurrences
    \item Normalization: Lowercase conversion
    \item Pre-tokenization: Whitespace splitting
\end{itemize}

\textbf{Sequence Processing:} Tokenized sequences are processed as follows:
\begin{enumerate}
    \item Truncation to maximum sequence length (1,028 tokens)
    \item Token ID remapping: $\text{remapped\_id} = \text{token\_id} \bmod \text{vocab\_size}$
    \item Padding with zero tokens to ensure uniform sequence length
\end{enumerate}

This tokenization strategy ensures that the model learns representations specifically adapted to tabular data characteristics while maintaining compatibility with transformer architectures. 

\begin{example}
\textbf{Illustrative Example for Tokenization and Augmentation.}

Consider the following toy dataset:

\begin{table}[h]
\centering
\begin{tabular}{l l l}
\toprule
\textbf{id} & \textbf{name} & \textbf{score} \\
\midrule
1 & Alice & 83 \\
2 & Bob   & 91 \\
\bottomrule
\end{tabular}
\caption{Dataset Version v1.}
\end{table}

A later version introduces a mild schema change and updated values:

\begin{table}[h]
\centering
\begin{tabular}{l l l}
\toprule
\textbf{id} & \textbf{full\_name} & \textbf{score} \\
\midrule
1 & Alice T. & 84 \\
2 & Bob      & 91 \\
\bottomrule
\end{tabular}
\caption{Dataset Version v2.}
\end{table}
\twocolumn
\paragraph{Linearization.}
Each table is converted into a flat sequence using structural tokens:

\begin{verbatim}
[COL] id [COL] name [ROW] 1 Alice [ROW] 2 Bob [VAL] 83 91
[COL] id [COL] full_name [ROW] 1 Alice T. [ROW] 2 Bob [VAL] 84 91
\end{verbatim}

\paragraph{Tokenization.}
Strings are first mapped to integer token IDs:
\begin{verbatim}
"id" → 17          "full_name" → 44
"Alice" → 5021     "Alice T." → 7422
"Bob" → 3195
\end{verbatim}

\paragraph{Vocabulary-Free Remapping.}
Given a fixed model vocabulary size $v$, token IDs are remapped via:
\[
\text{remapped\_id} = \text{token\_id} \bmod v .
\]

Applied to the values above:
\begin{verbatim}
5021 mod 5000 = 21
7422 mod 5000 = 2422
3195 mod 5000 = 3195
\end{verbatim}

\paragraph{Padding and Truncation.}
The resulting sequence is truncated to 1,028 tokens (discovered though ablation) and padded with zeros.
\paragraph{Outcome.}
Because both versions use the same structural linearization,
version-specific changes (e.g., \texttt{name}→\texttt{full\_name},  \\
\texttt{Alice}→\texttt{Alice T.}, or 83→84) translate into small but 
consistent differences in the token sequence, enabling the model to learn
a representation sensitive to version evolution.
\end{example}

\subsection{NT-Xent Contrastive Loss}

The core of our methodology employs the Normalized Temperature-scaled Cross Entropy (NT-Xent) loss, specifically adapted for table version discovery. Our design follows the SimCLR framework but incorporates domain-specific optimizations for tabular data.

Given two augmented views $z_i, z_j \in \mathbb{R}^{N \times d_{emb}}$ from the same batch of $N$ tables, our loss function operates as follows:

\textbf{Normalization:} We apply L2 normalization to ensure embeddings lie on the unit hypersphere in order to ensure values lie in the range of $[0,1]$:
$$z_i' = \frac{z_i}{||z_i||_2}, \quad z_j' = \frac{z_j}{||z_j||_2}$$

\textbf{Concatenation:} We create a combined representation matrix:
$$r = \begin{bmatrix} z_1' \\ \vdots \\ z_N' \\ z_1'' \\ \vdots \\ z_N'' \end{bmatrix} \in \mathbb{R}^{2N \times d_{emb}}$$

\textbf{Similarity Matrix:} We compute temperature-scaled cosine similarities:
$$\Theta = \frac{r \cdot r^T}{\tau} \in \mathbb{R}^{2N \times 2N}$$
where $\tau = 0.7$ is our temperature parameter, as per Fan, et. al \cite{fan2023semanticsawaredatasetdiscoverydata}

\textbf{Positive Pair Identification:} For each sample $i \in [0, N-1]$, its positive counterpart is located at index $i+N$:
$$\text{positive}(i) = \begin{cases} 
i + N & \text{if } i \in [0, N-1] \\
i - N & \text{if } i \in [N, 2N-1]
\end{cases}$$

\textbf{Loss Computation:} We apply the NT-Xent loss with self-similarity masking:
$$\mathcal{L} = -\frac{1}{2N} \sum_{i=0}^{2N-1} \log \frac{\exp(\Theta_{i,\text{positive}(i)})}{\sum_{j=0, j \neq i}^{2N-1} \exp(\Theta_{i,j})}$$

This formulation ensures that positive pairs (augmented views of the same table) are pulled together in the embedding space while being pushed away from all other tables in the batch, enabling effective version relationship learning. This also ensures that perturbations which can occur naturally (such as those we established earlier) can be learned, as they are closed in embedding space than those that cannot be. Following the analysis of Wang and Isola \cite{wang2022understandingcontrastiverepresentationlearning}, our contrastive objective can be inferred to perform a form of spectral clustering on the table similarity graph. Let $G = (V, E, W)$ be a weighted graph where $V$ represents tables, $E$ contains edges between semantically similar tables, and $W$ contains similarity weights.

The NT-Xent loss can be viewed as optimizing:
$$\min_f \mathbb{E}_{(T_i, T_j) \sim \mathcal{D}} \left[ -\log \frac{\exp(f(T_i)^T f(T_j) / \tau)}{\sum_{k \neq i} \exp(f(T_i)^T f(T_k) / \tau)} \right]$$

Under mild conditions, this objective converges to representations that preserve the dominant eigenvectors of the normalized graph Laplacian $L = I - D^{-1/2}WD^{-1/2}$, where $D$ is the degree matrix\cite{868688}. This convergence implicitly preserves semantics via spectral clustering, providing a basis for our contrastive method that can serve to provide a powerful embedding space even with minimal amounts of labeled data.

\subsection{SAVeD Architecture} 
SAVeD employs a custom transformer-based encoder specifically designed for table representation learning. Our architecture consists of the following components:

\textbf{Embedding Layer:} Input tokens are mapped to dense representations via a trainable embedding matrix $E \in \mathbb{R}^{v \times d_{model}}$, where $v$ is the vocabulary size and $d_{model}$ is the model dimension.

\textbf{Transformer Encoder:} We implement a multi-layer transformer encoder with $N$ layers, each containing multi-head self-attention with $h$ heads and position-wise feedforward networks. Following Vaswani et al.~\cite{DBLP:journals/corr/VaswaniSPUJGKP17}, we include residual connections and layer normalization.

\textbf{Pooling Strategy:} We apply mean pooling across the sequence dimension to obtain a fixed-size representation $H_{pool} \in \mathbb{R}^{B \times d_{model}}$, where $B$ is the batch size.

\textbf{Projection Head:} A two-layer feedforward network maps the pooled representation to the final embedding space: $Z = W_{out} \cdot \text{ReLU}(W_{in} \cdot H_{pool} + b_{in}) + b_{out}$, where $Z \in \mathbb{R}^{B \times d_{emb}}$.

\textbf{Regularization:} To address overfitting concerns identified in our preliminary experiments, we incorporate dropout layers with tunable rates in both transformer blocks and feedforward layers, L2 weight regularization with optimized decay parameter, and early stopping based on validation loss plateaus. 
Our architecture is fully differentiable and optimized end-to-end using the NT-Xent contrastive loss described above.

\section{Experimental Setup}

\subsection{Datasets and Benchmark}

We evaluate SAVeD on five canonical datasets from the Semantic Versioning in Databases (SDVB) Benchmark : IMDB, IRIS, NBA, TITANIC, and WINE\_small. 
Each dataset contains multiple version candidates created through realistic transformations including feature engineering, data cleaning, and preprocessing operations. It is important to note that, in the case of the SDVB benchmark, the authors provide an explicit list of version mappings, as well as which transformations were used to create the benchmark tables. Not all of the tables in SDVB that share a common root are versions of each other, i.e. \texttt{IMDB\_42} is not necessarily a version of \texttt{IMDB\_56}. 
As such, it is used to see whether SAVeD can adapt to external transformations as well, purely off the strength of the Transformer backbone architecture. 

\subsection{Evaluation Methodology}

Our evaluation framework implements a comprehensive assessment strategy that separates corpus-level analysis from pairwise classification tasks.
We implement a systematic evaluation across the entire dataset collection:
\begin{enumerate}
    \item \textbf{Embedding Generation:} All tables are embedded using the trained model to create a dense representation matrix $X \in \mathbb{R}^{N \times d_{emb}}$
    \item \textbf{Similarity Computation:} Pairwise cosine similarities are computed efficiently using normalized embeddings: $\theta(i,j) = \frac{X_i \cdot X_j}{||X_i|| \cdot ||X_j||}$
    \item \textbf{Category Analysis:} Similarities are categorized into:
    \begin{itemize}
        \item Self-similarities (diagonal elements, pair-pair matches)
        \item Intra-dataset similarities (same source dataset)
        \item Inter-dataset similarities (different source datasets)
    \end{itemize}
\end{enumerate}

\textbf{Evaluation Metrics:}
\\
\textbf{True Positive Rate and True Negative Rate}

We define the True Positive Rate (TPR) as rate of tables properly assigned to defined, in-datset versions (\texttt{Titanic 1 \& Titanic 4}, for example would contribute to this) over our threshold $\xi$. We also define true negatives tables that have similarity under the threshold $\xi$, while coming from another source dataset. The intution is that optimizing these metrics will allow us to differentiate between source dataset semantics and pre-emptively lower the search space for candidates when searching for versions of an unseen dataset. 
\\
\textbf{Separation Score:} Measures the model's ability to distinguish between version and non-version pairs. We define separation as:
$$\text{Separation} = \mu(\theta(\text{intra-dataset})) - \mu(\theta(\text{inter-dataset}))$$
where $\mu(\theta)$ is a mean of the similarity across an entire dataset. 




\subsection{Training}
We trained with the AdamW optimizer \cite{Loshchilov2017DecoupledWD} with the a learning rate of $2.3 \times 10^{-4}$ (optimized via hyperparameter search), weight decay of $5.7 \times 10^{-5}$ for L2 regularization, beta parameters $\beta_1 = 0.9$ and  $\beta_2 = 0.999$, and a batch size of 32, which is chosen for a balance of memory efficiency and gradient stability). Our training protocol involves a linear learning rate warm-up over first 10\% of training steps, followed by the main contrastive learning process. To mitigate overfitting, we add patience-based early stopping which monitors changes in the validation loss. Our optimal model is selected on the minimization of validation loss and maximization of separation. For cross-validation, we implement a stratified splitting approach, where 70\% of tables per dataset are used for training, 15\% of the tables per dataset are used for for hyperparameter optimization and early stopping, and the final 15\% are used for the final evaluation and are completely held out during training.
Each training batch contains tables from multiple datasets to encourage generalization. Positive pairs are constructed within-batch through augmentation, while negative pairs are formed from all other tables in the batch.
We also implement gradient clipping (max norm = 1.0) to prevent exploding gradients during contrastive learning, particularly important given the temperature scaling in our loss function. 
\subsection{Baseline Comparisons}
Since there are currently no existing version discovery methods, we compare SAVeD against a comprehensive set of representative methods for related table and table union search, which we consider to be the closest task. For LLMs, we use the methodology of DiscoverGPT \cite{hu-etal-2025-discovergpt} for comparing version pair candidates. DiscoverGPT builds upon work from DataLore \cite{Lou2024}, aiming to be a cross-task methodology utilizing LLMs for database tasks. Intuitively, DiscoverGPT's should perform well on the task of version discovery, dependent on the LLM engine used for evaluation.

\begin{itemize}
    \item \textbf{ Starmie\cite{fan2023semanticsawaredatasetdiscoverydata}:} State-of-the-art pre-trained RoBERTa-based table understanding model\cite{10.14778/3407790.3407797} adapted for version discovery through supervised fine-tuning
    \item \textbf{DiscoverGPT (Llama2-7B Backbone)} \cite{touvron2023llama2openfoundation}- This uses the methodology of DiscoverGPT with the LLama2-7B LLM.
    \item \textbf{DiscoverGPT (Qwen2.5-7B Backbone) :}\cite{qwen2025qwen25technicalreport}- This uses the methodology of DiscoverGPT with the Qwen2.5-7B LLM.
    \item \textbf{DiscoverGPT (Gemma2-7B Backbone) :} \cite{gemmateam2024gemma2improvingopen}-This uses the methodology of DiscoverGPT with the Gemma2-7B LLM.
\end{itemize}

\subsection{Design Details}

All experiments use a fixed random seed with deterministic CUDA \cite{cuda} operations enabled to ensure reproducible results across runs. Our design features comprehensive seed setting for Python's random module \cite{van1995python}, NumPy\cite{harris2020array}, and PyTorch's CPU and GPU random number generators \cite{DBLP:journals/corr/abs-1912-01703}. All experiments were run on an NVIDIA RTX 3090 using CUDA version 12.9. Flash Attention 2 \cite{dao2023flashattention2fasterattentionbetter} was used for faster inference where possible. 

\section{Results}

\subsection{Main Results}

Table~\ref{tab:results} presents our main experimental results across all five benchmark datasets. SAVeD demonstrates strong performance, achieving the highest true positive rate on 3 out of 5 datasets and competitive results on the remaining two. We use the inbuilt Pytorch method for determining parameter counts for models.

\section{Conclusion}

We present SAVeD (Semantically Aware Version Discovery), a novel contrastive learning framework that addresses the critical but understudied problem of automated version discovery in data lakes. Our approach demonstrates that semantic version relationships can be effectively learned without explicit supervision through carefully designed table augmentations and transformer-based representation learning.

\newpage
\clearpage
\bibliographystyle{ACM-Reference-Format}
\bibliography{sample}
\clearpage
\onecolumn
\FloatBarrier
\appendix
\section*{Appendix 1: List of Symbols and Variables}
\begin{table*}[!h]
    \caption{Variables and What They Mean}
    \label{tab:names}
    \centering
    \begin{tabular}{|c|c|}
        	\toprule
        	\text{Notation}& \text{Description} \\
        \midrule
         $T$ & \text{An original table}\\
         $T'$ & \text{A version of $T$}\\
         $T_{ori}$ & \text{Unaugmented table created from the positive sample of the batch of the table}\\
         $T_{aug}$ & \text{Augmented table created from the positive sample of the batch of the table}\\
         $N$ & \text{Batch Size}\\
         $n$ & \text{Sequence Length}\\
         $v$ & \text{Vocabulary Size}\\
         $d_{model}$ & \text{Hidden Dimension of the Model}\\
         $d_{emb}$ & \text{Output Embedding Dimension}\\
         $d_{hidden}$ & \text{Weight Hidden Dimension}\\
         $h$ & \text{Number of heads used for attention} \\
         $L$ & \text{Number of transformer layers}\\
         $\mathbb{X}$ & \text{Input Tensor $\in \mathbb{R}^{b\times n}$}\\
         $E$ &\text{Trainable Embedding Matrix $\in \mathbb{R}^{v\times d_{model}}$}\\
         $X_{emb}$ & \text{Embedded Input $\in \mathbb{R} ^ {B \times n \times d_{model}}$}\\
         $H$ & \text{Table Encoder Output $\in \mathbb{R} ^{B \times n \times d_{model}}$}\\
         $H_{pool}$ & \text{Spatially reduced representation of the Encoder output $\in \mathbb{R}^{b\times d_{model}}$}\\
         $Z$ & \text{Output Embedding $\in \mathbb{R} ^ {b \times d_{emb}}$}\\
         $W_{in}$ & \text{Input weight matrix $\in \mathbb{R} ^ {d \times d_{hidden}}$} \\
         $W_{out} $ & \text{Output weight matrix $\in \mathbb{R}^{d_{hidden \times d_{emb}}}$}\\
         $\beta$ & \text{Trainable shift parameter in BatchNorm}\\
         $\gamma$ & \text{Trainable scale parameter in BatchNorm}\\
         $\Theta$ & \text{similarity matrix}\\
         $\theta$ & \text{similarity metric}\\
         $\mathcal{L}$ & \text{Loss}\\
         $\tau$& \text{temperature parameter that controls sharpness of distribution}\\
         $\mathcal{M}$ & \text{mask applied to remove self-similarity in loss function $\in \mathbb{R}^{2N\times 2N}$}\\
         $\mathbb{L}$ & \text{label matrix created in loss function $\in \mathbb{R}^{2N\times 2N}$}\\
         $r$ & \text{concatenation of normalized embeddings for all samples in a batch $\in \mathbb{R}^{2N\times d_{emb}}$}\\
        \bottomrule
    \end{tabular}
\end{table*}

\clearpage

\section*{Appendix 2: DiscoverGPT Prompt}
\begin{lstlisting}
Compare these two tables and determine if they represent different
semantic versions of the same dataset:

Table 1:
{table1_text}

Table 2:
{table2_text}

Analysis: These tables are {"versions" if "similar structure"
in table1_text.lower() and "similar structure" in
table2_text.lower() else "non-versions"}
\end{lstlisting}

\vspace{0.5pt}

\section*{Appendix 3: Titanic Dataset Transformations}
\begin{table}[!htbp]
    \centering
    \captionsetup{font=small}
    \small
    \resizebox{\columnwidth}{!}{%
    \begin{tabular}{|c|c|c|c|c|c|c|c|c|c|c|c|}
    \hline
        	extbf{PassengerId} & \textbf{Survived} & \textbf{Pclass} & \textbf{Name} & \textbf{Sex} & \textbf{Age} & \textbf{SibSp} & \textbf{Parch} & \textbf{Ticket} & \textbf{Fare} & \textbf{Cabin} & \textbf{Embarked} \\
        \hline
        1 & 0 & 3 & Braund, Mr. Owen Harris & male & 22.0 & 1 & 0 & A/5 21171 & 7.25 & NaN & S \\
        \hline
        2&1&1&Cumings, Mrs. John Bradley (Florence Briggs Thayer)&female&38.0&1&0&PC 17599&71.2833&C85&C\\
        \hline
        3&1&3&Heikkinen, Miss. Laina&female&26.0&0&0&STON/O2. 3101282&7.92560&NaN&S\\
        \hline
        4&1&1&Futrelle, Mrs. Jacques Heath (Lily May Peel)&female&35.0&1&0&113803&53.1000&C123&S\\
        \hline
        5&0&3&Allen, Mr. William Henry&male&35.0&0&0&373450&8.0500&NaN&S\\
        \hline
    \end{tabular}
    }
    \caption{Base Titanic Dataset (12 Columns)}
    \label{tab:basetitanic}
\end{table}

\begin{table}[!htbp]
    \centering
    \captionsetup{font=small}
    \small
    \resizebox{\columnwidth}{!}{%
    \begin{tabular}{|c|c|c|c|c|c|c|c|c|c|c|c|c|c|}
    \hline
        	extbf{PassengerId} & \textbf{Survived} & \textbf{Pclass} & \textbf{Age} & \textbf{SibSp} & \textbf{Parch} & \textbf{Fare} & \textbf{Name\_Abbing, Mr. Anthony} & \textbf{Name\_Abbott, Mr. Rossmore Edward} & \textbf{Name\_Abbott, Mrs. Stanton (Rosa Hunt)} & \dots & \textbf{Cabin\_F33} & \dots & \textbf{Embarked\_nan} \\
        \hline
        1 & 0 & 3 & 22.0 & 1 & 0 & 7.2500 & 0.0 & 0.0 & 0.0 & \dots & 0.0 & \dots & 0.0 \\
        \hline
        2 & 1 & 1 & 38.0 & 1 & 0 & 71.2833 & 0.0 & 0.0 & 0.0 & \dots & 0.0 & \dots & 0.0 \\
        \hline
        3 & 1 & 3 & 26.0 & 0 & 0 & 7.9250 & 0.0 & 0.0 & 0.0 & \dots & 0.0 & \dots & 0.0 \\
        \hline
        4 & 1 & 1 & 35.0 & 1 & 0 & 53.1000 & 0.0 & 0.0 & 0.0 & \dots & 0.0 & \dots & 0.0 \\
        \hline
        5 & 0 & 3 & 35.0 & 0 & 0 & 8.0500 & 0.0 & 0.0 & 0.0 & \dots & 0.0 & \dots & 0.0 \\
        \hline
    \end{tabular}
    }
    \caption{One-Hot Encoded Titanic Dataset (1733 Columns)}
    \label{tab:ohe_titanic}
\end{table}

\begin{table}[!htbp]
    \centering
    \captionsetup{font=small}
    \small
    \resizebox{\columnwidth}{!}{%
    \begin{tabular}{|c|c|c|c|c|c|c|c|c|c|c|c|c|c|}
    \hline
        	extbf{PassengerId} & \textbf{Survived} & \textbf{Pclass} & \textbf{Name} & \textbf{Sex} & \textbf{Age} & \textbf{SibSp} & \textbf{Parch} & \textbf{Ticket} & \textbf{Fare} & \dots & \textbf{Cabin\_emb\_43} & \dots & \textbf{Embarked\_emb\_2} \\
        \hline
        1 & 0 & 3 & 109 & 2 & 22.0 & 1 & 0 & 524 & 7.2500 & \dots & 0.427617 & \dots & -0.173057 \\
        \hline
        2 & 1 & 1 & 191 & 1 & 38.0 & 1 & 0 & 597 & 71.2833 & \dots & 0.125830 & \dots & -0.997602 \\
        \hline
        3 & 1 & 3 & 354 & 1 & 26.0 & 0 & 0 & 670 & 7.9250 & \dots & 0.427617 & \dots & -0.173057 \\
        \hline
        4 & 1 & 1 & 273 & 1 & 35.0 & 1 & 0 & 50 & 53.1000 & \dots & -0.727363 & \dots & -0.173057 \\
        \hline
        5 & 0 & 3 & 16 & 2 & 35.0 & 0 & 0 & 473 & 8.0500 & \dots & 0.427617 & \dots & -0.173057 \\
        \hline
    \end{tabular}
    }
    \caption{Titanic Dataset with Embeddings (167 columns)}
    \label{tab:embed_titanic}
\end{table}

\section*{Appendix 4: Results}
\begin{table}[h]
    \centering
    \caption{Performance Comparison on Version Discovery Task}
    \label{tab:results}
    \renewcommand{\arraystretch}{1.0}
    \small
    \resizebox{\columnwidth}{!}{%
    \begin{tabular}{lcccccc}
        	\toprule
        	\textbf{Method} &
        	\textbf{IMDB(TPR/Separation)} &
        	\textbf{IRIS(TPR/Separation)} &
        	\textbf{NBA(TPR/Separation)} &
        	\textbf{TITANIC(TPR/Separation)} &
        	\textbf{WINE\_small (TPR/Separation)} & 
            \textbf{Number of Parameters} \\
        \midrule
        
        Starmie & 17.12 / 0.0022 & 17.12 / 0.0030 & 16.63 / 0.0019 & 19.60 / 0.0018 & 28.54 / 0.0022 & 124M\\
        Discover-GPT (Qwen2.5-7B) & 74.06 / 0.0693 & 50.00 / 0.1738 & 47.08 / -0.0708 & 48.61 / 0.1353 & 68.25 / 0.1226&7B \\
        Discover-GPT (Gemma-7B) & 78.44 / 0.3018 & 75.94 / -0.0972 & 85.83 / 0.1394 & 81.94 / 0.4009 & 83.25 / 0.3887&7B \\
        Discover-GPT (LLaMA2-7B) & 80.94 / 0.1123 & \textbf{94.37} / 0.4871 & 90.42 / 0.1304 & \textbf{97.22} / 0.3696 & 85.50 / 0.4917 &7B\\
        	\textbf{SAVeD (Ours)} & \textbf{88.75} / \textbf{0.3916} & 90.62 / \textbf{0.8010} & \textbf{90.83} / \textbf{0.4442} & 88.89 / \textbf{0.7717} & \textbf{87.00} / \textbf{0.5342} &14M\\
        \bottomrule
    \end{tabular}
    }
\end{table}
\end{document}